\documentclass[letterpaper, 10 pt, conference]{ieeeconf}

\IEEEoverridecommandlockouts
\usepackage{graphicx}
\usepackage{amsmath,amssymb}
\usepackage{amsfonts}
\usepackage{booktabs}
\usepackage{xcolor}
\usepackage{caption}
\usepackage{subcaption}
\usepackage{algorithm}
\usepackage{algpseudocode}
\usepackage{tikz}
\usetikzlibrary{arrows.meta,positioning,calc,fit,backgrounds,shapes.geometric,3d}
\usepackage[colorlinks=true,allcolors=blue]{hyperref}
\usepackage{cite}
\usepackage{microtype}

\newcommand{\clipseg}{\mbox{CLIPSeg}}
\newcommand{\vct}[1]{\mathbf{#1}}
\newcommand{\norm}[1]{\lVert #1 \rVert}

\title{\LARGE \bf
Towards Spatial Perception for Heterogeneous Robot Collaboration\\
in Subterranean Mining Environments}

\author{Mario A.V. Saucedo, Akash Patel, 
Christoforos Kanellakis, and George Nikolakopoulos%
\thanks{This work has received funding from the European Union's Horizon Europe
Research and Innovation Program, under the Grant Agreement No.~101138451
(PERSEPHONE), the European Regional Development Fund and the Campus Totalförsvar Övre Norrland project (No. 20375828).}%
\thanks{The authors are with the Robotics and Artificial Intelligence Group,
Department of Computer Science, Electrical and Space Engineering,
Lule{\aa} University of Technology, 971~87 Lule{\aa}, Sweden.
{Corresponding author`s email: \tt marval@ltu.se}}%
}

\begin{document}

\maketitle
\thispagestyle{empty}
\pagestyle{empty}

\begin{abstract}
The autonomous extraction of deep mineral deposits in abandoned underground mines
is fundamentally a multi-agent integration problem. No single platform
simultaneously offers the mobility to traverse kilometers of degraded drifts and
the sensing payload required to characterize an ore body. This article presents
the onboard perception pipeline that bridges two heterogeneous agents within the
PERSEPHONE autonomous mining mission. Which consist of a lightweight \emph{Explorer} robot that maps an
unknown mine and generates a 3D scene graph of inspection targets, 
by running a zero-shot, vision-language semantic segmentation stack
that detects mineral deposits directly from natural-language prompts. 
The map and the graph are then handed to a
second \emph{Inspector} robot, which carries an advanced sensing payload and uses
them to plan close-range inspection viewpoints. We detail the complete pipeline, with
emphasis on the geometric abstraction that turns raw detections into actionable
inspection targets, spanning per-view bounding-box generation, cross-view box merging,
plane fitting, and polygon extraction, and we report an extensive field validation in a
subterranean test facility and in an active magnesite mine, covering both iron-vein
and magnesite mineralization under realistic, perceptually degraded conditions.
\end{abstract}

\section{Introduction}\label{sec:intro}

The transition towards a low-carbon economy is sharply increasing the demand for
critical raw materials, renewing interest in the vast resources locked inside
abandoned or partially exploited underground mines. Re-entering these sites with
human crews is slow, costly, and dangerous. The drifts are unlit, unventilated, and
structurally uncertain, and the location and grade of any remaining mineralization are
largely unknown. Autonomous robots are therefore an attractive enabling technology,
and recent subterranean robotics deployments have shown that mobile platforms can map
and traverse such environments with a high degree of autonomy~\cite{tranzatto2022cerberus,
agha2021nebula,hudson2022csiro,dang2020graph}.

A recurring lesson from these efforts is that no single robot is simultaneously
optimal for exploring a large unknown network and for characterizing
a specific resource at close range. Long-range exploration favors an agile, lightly
instrumented platform that can cover distance and build a map quickly. In contrast,
resource characterization favors a slower, heavily instrumented platform carrying,
for example, a hyperspectral camera, a ground-penetrating sensor, or a drilling tool
whose payload and power budget make broad exploration impractical. The natural
architecture is therefore heterogeneous. An \emph{Explorer} discovers and
coarsely localize the resource, and an \emph{Inspector} revisits the
most promising regions to perform high-value, close-range sensing. This division of
labor underpins the PERSEPHONE\footnote{PERSEPHONE: Autonomous Exploration and Extraction of
Deep Mineral Deposits. \url{https://www.persephone-mining.eu/}} mission and is the
context for the present work.

\begin{figure}[t]
  \centering
  \includegraphics[width=\linewidth]{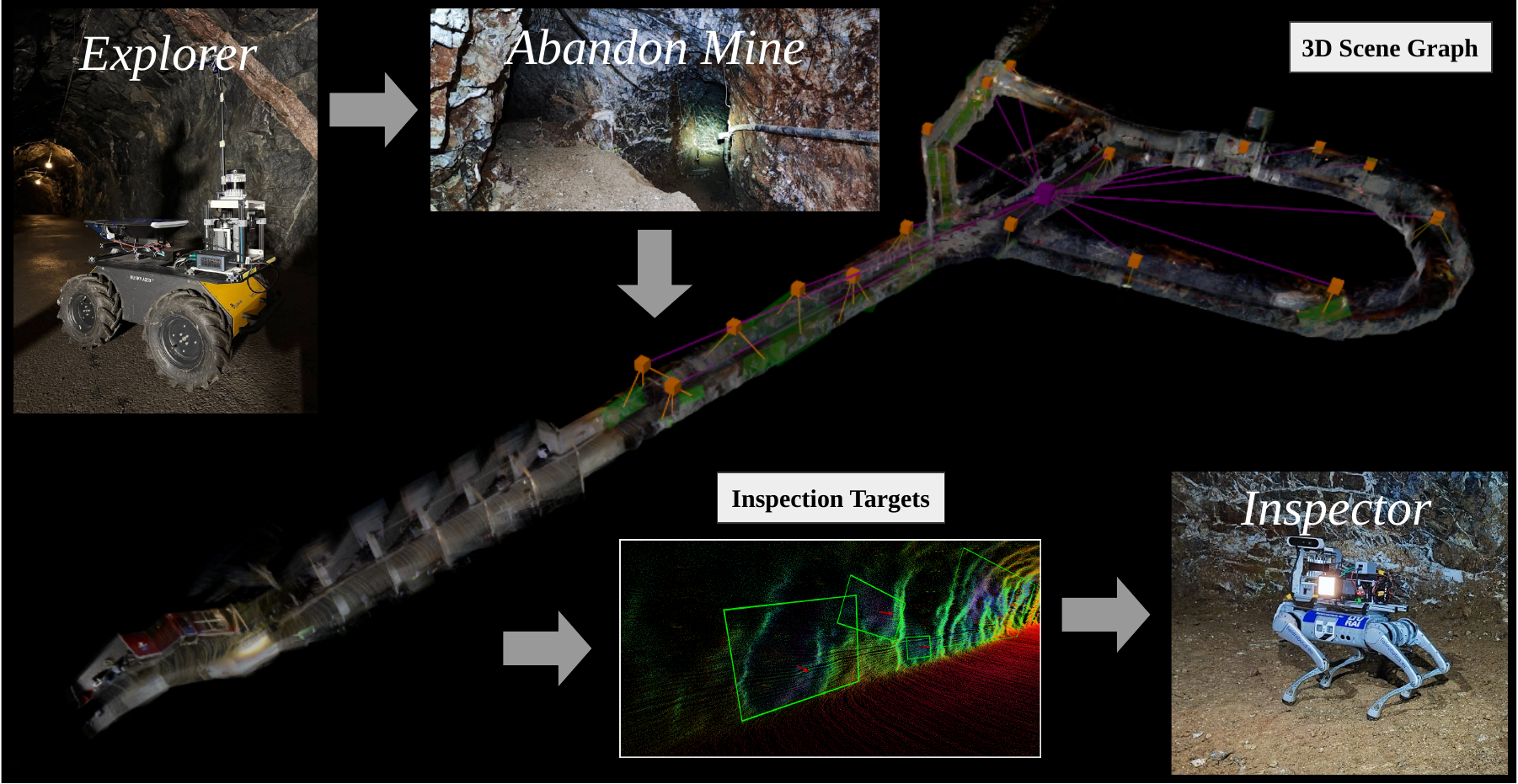}
  \caption{\textbf{Heterogeneous two-robot mining mission.} The \emph{Explorer} maps the mine while
  running the proposed onboard open-set mineral perception pipeline, it then transmits a PCD map and
  a set of oriented mineral polygons to the \emph{Inspector}, which plans close-range
  inspection along each polygon's normal.}
  \label{fig:system}
\end{figure}

The challenge of such a collaboration is not the locomotion of either robot in isolation,
but the interface between them. For the Inspector to act on the Explorer's
findings, the Explorer must hand over more than a point-cloud map. It must communicate
where the mineralization is, what it is, and how a downstream
robot should position itself to observe it. The handover must be compact enough to be
transmitted over an intermittent underground link, geometrically expressive enough to
drive a viewpoint planner, and produced fully onboard, in real time, without any
prior model of the specific mine. This article focuses on the perception pipeline that
realize exactly this interface.

Two requirements make the perception problem distinctive. First, the targets are
defined semantically and openly. An operator may ask for ``magnesite'' in one deployment and ``iron vein'' in another, and collecting and
annotating a representative training set for each mineral, mine, and lighting
condition is impractical, hazardous, and expensive. We therefore adopt a
zero-shot vision-language segmentation model, \clipseg{}~\cite{luddecke2022clipseg},
which segments image regions from a free-text prompt without any task-specific
training, and we ground its predictions in 3D through LiDAR projection. Second, the
output must be actionable. A heat-map of mineral probability over a point cloud
is not directly serviceable by an inspection planner. The pipeline must then distill the
dense, noisy, multi-view evidence into a small number of discrete, well-posed
inspection targets.

We close this gap with a geometric abstraction that, to the best of our knowledge,
has not previously been used for cross-robot mineral handover. For each detected
deposit we accumulate the mineral-labeled points across viewpoints, cluster them into
a persistent instance, fit a planar surface to the cluster, project those points onto
the fitted plane, and extract their minimum-area bounding rectangle. The
result is an oriented polygon that simultaneously encodes the spatial extent of
the mineralize patch and the surface normal along which it should be inspected.
These polygons are not handled in isolation. Building on our open-set perception stack
and on a semantically driven clustering~\cite{saucedo2026areas},
the polygons are organized into an underlying 3D scene graph of inspection targets. This lets the
mission reason about the deposits globally rather than one at a time, such that more densely
accumulated mineralization can be prioritized for inspection first.

The contributions of this article are:
\begin{itemize}
  \item A complete, onboard, open-set mineral perception pipeline that fuses
        zero-shot vision-language segmentation with LiDAR to detect and localize
        mineral deposits in 3D, using only natural-language prompts and no
        mine-specific training data.
  \item A geometric abstraction that turns noisy, redundant per-frame detections into
        actionable inspection targets: cross-view association condenses them into
        persistent, instance-level regions of interest, each region is reduced to an
        oriented polygon by fitting a plane to its mineral points and extracting the
        minimum-area bounding rectangle of their projections onto that plane, and the
        polygons are organized into a 3D scene graph. 
        Together these constitute the Explorer-to-Inspector interface.
  \item An extensive experimental validation across two qualitatively different
        environments (a subterranean test facility and an active magnesite mine) and
        two mineralization types (iron veins and magnesite), under realistic
        underground conditions.
\end{itemize}


\begin{figure*}[t]
  \centering
    \resizebox{\linewidth}{!}{%
  \begin{tikzpicture}[
    font=\normalsize,
    arr/.style={-{Stealth[length=5.5mm,width=3.2mm]}, line width=3.2pt, black!28},
    iarr/.style={-{Stealth[length=3mm,width=1.8mm]}, line width=1.4pt, black!50},
    lbl/.style={font=\scriptsize\bfseries, text=black!60, fill=white,
                inner sep=2pt, rounded corners=2pt, align=center},
    inp/.style={draw=gray!45, fill=gray!10, rounded corners=5pt,
                minimum width=28mm, minimum height=9mm,
                align=center, line width=0.7pt, font=\small\bfseries},
  ]

  \definecolor{dksteel}{RGB}{36,41,46}

  \node[fill=dksteel, fill opacity=0.88, rounded corners=5pt,
        minimum width=26mm, minimum height=28mm] (seg) at (  0mm,0) {};
  \node[fill=dksteel, fill opacity=0.88, rounded corners=5pt,
        minimum width=26mm, minimum height=28mm] (fus) at ( 50mm,0) {};
  \node[fill=dksteel, fill opacity=0.88, rounded corners=5pt,
        minimum width=26mm, minimum height=28mm] (clu) at (100mm,0) {};
  \node[fill=dksteel, fill opacity=0.88, rounded corners=5pt,
        minimum width=26mm, minimum height=28mm] (geo) at (150mm,0) {};
  \node[fill=dksteel, fill opacity=0.88, rounded corners=5pt,
        minimum width=26mm, minimum height=28mm] (out) at (200mm,0) {};

  \begin{scope}[shift={(0mm,6mm)}]
    \fill[white, fill opacity=0.18]
      (-7mm,0) .. controls (-3mm,5mm) and (3mm,5mm) .. (7mm,0)
               .. controls (3mm,-5mm) and (-3mm,-5mm) .. (-7mm,0);
    \draw[white, line width=1.4pt]
      (-7mm,0) .. controls (-3mm,5mm) and (3mm,5mm) .. (7mm,0)
               .. controls (3mm,-5mm) and (-3mm,-5mm) .. (-7mm,0);
    \fill[white, fill opacity=0.88] (0,0) circle (2.5mm);
    \fill[black!65]                 (0,0) circle (1.4mm);
    \fill[white]                    (0.7mm,0.7mm) circle (0.5mm);
  \end{scope}

  \begin{scope}[shift={(50mm,6mm)}]
    \draw[white!75, line width=0.9pt, -{Stealth[length=1.8mm,width=1.1mm]}]
      (0,0) -- (0,6mm);
    \draw[white!75, line width=0.9pt, -{Stealth[length=1.8mm,width=1.1mm]}]
      (0,0) -- (7mm,-1mm);
    \draw[white!75, line width=0.9pt, -{Stealth[length=1.8mm,width=1.1mm]}]
      (0,0) -- (-4mm,-3mm);
    \fill[white] (0,0) circle (0.45mm);
    \fill[white, fill opacity=0.45] ( 4mm, 1mm) circle (0.75mm);
    \fill[white, fill opacity=0.60] ( 3mm, 2mm) circle (0.75mm);
    \fill[white, fill opacity=0.40] (-1mm, 2mm) circle (0.75mm);
    \fill[white, fill opacity=0.55] ( 5mm, 3mm) circle (0.75mm);
    \fill[white, fill opacity=0.35] ( 2mm, 1mm) circle (0.75mm);
    \fill[white, fill opacity=0.65] ( 3mm, 4mm) circle (0.75mm);
    \fill[yellow!90] (3mm,4mm) circle (1.3mm);
  \end{scope}

  \begin{scope}[shift={(100mm,6mm)}]
    \draw[white, fill=white, fill opacity=0.10,
          line width=0.7pt, rounded corners=1mm]
      (-7mm,-4mm) rectangle (0mm,3mm);
    \draw[white, fill=white, fill opacity=0.20,
          line width=0.8pt, rounded corners=1mm]
      (-4mm,-3mm) rectangle (4mm,4mm);
    \draw[white, fill=white, fill opacity=0.28,
          line width=0.9pt, rounded corners=1mm]
      (-1mm,-2mm) rectangle (6mm,5mm);
    \draw[yellow!88, line width=1.2pt, rounded corners=1mm]
      (-1mm,-2mm) rectangle (6mm,5mm);
  \end{scope}

  \begin{scope}[shift={(150mm,6mm)}]
    \fill[white, fill opacity=0.20]
      (-6mm,-3mm) -- (4mm,-5mm) -- (8mm,1mm) -- (-2mm,3mm) -- cycle;
    \draw[white, line width=1.2pt]
      (-6mm,-3mm) -- (4mm,-5mm) -- (8mm,1mm) -- (-2mm,3mm) -- cycle;
    \draw[yellow!88, line width=1.2pt,
          -{Stealth[length=2mm,width=1.3mm]}]
      (1mm,-1mm) -- (-0.5mm,5mm);
    \node[text=yellow!88, font=\scriptsize\bfseries] at (0mm,6.5mm) {$\hat{n}$};
  \end{scope}

  \begin{scope}[shift={(200mm,6mm)}]
    \draw[white!50, line width=0.8pt] (-5mm,1mm) -- (0mm,5mm);
    \draw[white!50, line width=0.8pt] (0mm,5mm)  -- (5mm,1mm);
    \draw[white!50, line width=0.8pt] (-5mm,1mm) -- (-1mm,-4mm);
    \draw[white!50, line width=0.8pt] (-1mm,-4mm) -- (5mm,1mm);
    \draw[white!50, line width=0.8pt] (-5mm,1mm) -- (5mm,1mm);
    \fill[yellow!90]  (0mm,5mm)  circle (2.6mm);
    \fill[white]      (-5mm,1mm) circle (1.9mm);
    \fill[white]      (5mm,1mm)  circle (1.9mm);
    \fill[white!55]   (-1mm,-4mm) circle (1.3mm);
  \end{scope}

  \node[text=white, align=center] at (  0mm,-5mm)
    {\textbf{\normalsize See}\\[0.8mm]
     {\scriptsize\color{yellow!80!white}Open-Set Detection}};

  \node[text=white, align=center] at ( 50mm,-5mm)
    {\textbf{\normalsize Locate}\\[0.8mm]
     {\scriptsize\color{yellow!80!white}3D Grounding}};

  \node[text=white, align=center] at (100mm,-5mm)
    {\textbf{\normalsize Track}\\[0.8mm]
     {\scriptsize\color{yellow!80!white}Multi-View Fusion}};

  \node[text=white, align=center] at (150mm,-5mm)
    {\textbf{\normalsize Model}\\[0.8mm]
     {\scriptsize\color{yellow!80!white}Planar Polygon}};

  \node[text=white, align=center] at (200mm,-5mm)
    {\textbf{\normalsize Plan}\\[0.8mm]
     {\scriptsize\color{yellow!80!white}Scene Graph}};

  \node[inp] (rgb) at ([xshift=-16mm,yshift=12mm]seg.north) {RGB Image};
  \node[inp] (txt) at ([xshift= 16mm,yshift=12mm]seg.north) {Language Prompt};
  \node[inp] (lid) at ([yshift=12mm]fus.north) {LiDAR Cloud};
  \node[inp] (odo) at ([yshift=12mm]clu.north) {Odometry};

  \draw[iarr] (rgb.south) -- ([xshift=-6mm]seg.north);
  \draw[iarr] (txt.south) -- ([xshift= 6mm]seg.north);
  \draw[iarr] (lid.south) -- (fus.north);
  \draw[iarr] (odo.south) -- (clu.north);

  \draw[arr] (seg.east) -- node[lbl,above=3pt]{labelled\\masks}      (fus.west);
  \draw[arr] (fus.east) -- node[lbl,above=3pt]{semantic\\3D cloud} (clu.west);
  \draw[arr] (clu.east) -- node[lbl,above=3pt]{deposit\\regions}    (geo.west);
  \draw[arr] (geo.east) -- node[lbl,above=3pt]{oriented\\polygons}  (out.west);

  \end{tikzpicture}%
  }
  \caption{\textbf{Onboard perception pipeline.} The \emph{Explorer} transforms its raw sensor
  streams (top: RGB image, language prompt, LiDAR cloud, and odometry) into a
  compact 3D scene graph of inspection targets through five stages.
  \textbf{See}~--~zero-shot segmentation with \clipseg{} detects the prompted mineral in
  the image; \textbf{Locate}~--~the labeled mask is fused with the synchronized LiDAR
  scan to ground each detection in 3D; \textbf{Track}~--~detections are merged and
  deduplicated across the trajectory into persistent deposit regions;
  \textbf{Model}~--~each region is abstracted as an oriented planar polygon encoding
  position, extent, and surface normal; \textbf{Plan}~--~the polygons are organized into
  a 
  3D scene graph and transmitted to the \emph{Inspector}. 
  }
  \label{fig:pipeline}
\end{figure*}
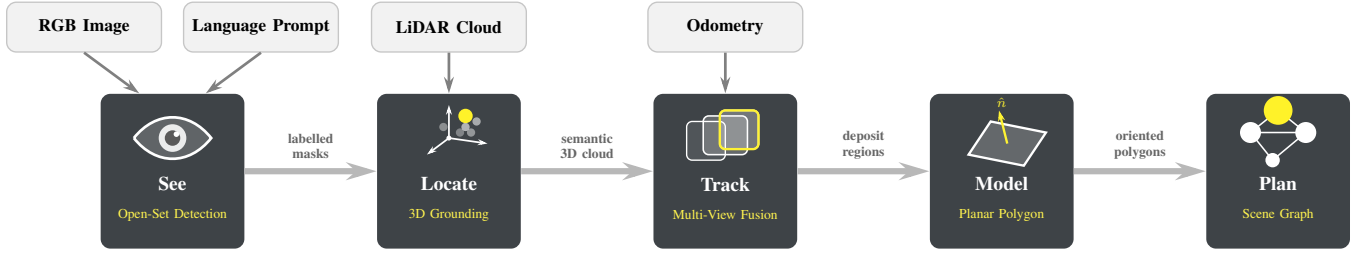

\section{Related Work}\label{sec:related}

\textbf{Autonomy in subterranean and mining environments.}
Robotic autonomy underground matured rapidly through the DARPA Subterranean Challenge,
which pushed teams to map and traverse tunnels, caves, and built substructures under the
communication and perception constraints that define such spaces. The leading entries,
namely CERBERUS~\cite{tranzatto2022cerberus}, Team CoSTAR's
NeBula~\cite{agha2021nebula}, CSIRO Data61~\cite{hudson2022csiro}, and
CTU-CRAS-NORLAB~\cite{rouvcek2021darpa}, converged on a common conclusion, that no single
morphology suffices and that heterogeneous fleets of legged, wheeled, and aerial robots
cover such environments far more effectively than any one platform. On this foundation,
exploration has become a reliable competence, ranging from graph-based
planners~\cite{dang2020graph} to traversability-aware variants such as
STAGE~\cite{patel2024stage} that give an Explorer its long-range mapping, and the broader
case for multi-robot autonomy in mining is by now equally well
argued~\cite{patel2025mining}. This body of work is concerned
primarily with reaching and covering the environment, through robust locomotion, state
estimation, and exploration, and it treats what it maps as geometry rather than as
content to be understood. The complementary problem we address is how to recognize a
specific, operator-defined resource within that geometry and package it so that a second,
specialized robot can act on it. This calls first for a perception front-end able to name
targets with no mine-specific training.

\textbf{Open-vocabulary and zero-shot visual perception.}
Such open-ended recognition has become possible only recently, through vision-language
models. CLIP~\cite{radford2021clip} aligns images and text in a shared embedding space
and in doing so recognizes categories never seen during fine-tuning, and a substantial
line of work has since carried this ability into dense prediction.
LSeg~\cite{li2022lseg} aligns per-pixel features with text embeddings, and
\clipseg{}~\cite{luddecke2022clipseg} attaches a lightweight decoder to a
frozen backbone to segment an image from an arbitrary text or image prompt. For a
field deployment that must respond to operator-specified minerals with no opportunity to
collect data, the training-free, prompt-driven behavior of \clipseg{} is the decisive
property, and we adopt it as our semantic front-end. Where most of these studies are
judged on curated benchmarks, our concern is their behavior on raw, low-light,
dust-laden mine imagery, and just as importantly the question of how to lift their 2D
predictions into a consistent 3D representation.

\textbf{Open-set 3D mapping and semantic scene understanding.}
Lifting language into 3D is by now an active field in its own right.
ConceptFusion~\cite{jatavallabhula2023conceptfusion} fuses pixel-aligned features into an
open-set 3D map, OpenScene~\cite{peng2023openscene} co-embeds 3D points with CLIP features
for open-vocabulary queries, and ConceptGraphs~\cite{gu2024conceptgraphs} assembles
open-vocabulary 3D scene graphs for planning. These representations are rich and
queryable, but they are also heavy to compute, store, and transmit, which is precisely
what a bandwidth-constrained, multi-robot mining mission cannot afford. Such a mission
needs the opposite, a compact, task-specific description that a downstream planner can
act on directly. We obtain it from a single observation, that individual semantic
entities can be clustered into coherent \emph{areas} on a 3D scene
graph~\cite{saucedo2026areas}, and we apply the same principle to aggregate mineral
detections into a small set of prioritizable inspection regions. This places our
contribution between low-level open-set perception and high-level mission planning over a
graph of inspection targets, and such targets are useful only once a planner can turn
them into motion.

\textbf{Viewpoint generation and inspection planning.}
Deciding where to place a sensor to observe a surface is the classical problem of view
planning and coverage. Receding-horizon next-best-view planning and structural inspection
path planning~\cite{bircher2016nbv,almadhoun2019survey}
generate viewpoints over a known or partially known surface, and the Inspector in the
proposed framework builds on an adaptive inspection planner~\cite{viswanathan2025adaptive}
that refines a prior view-plan against the surface it actually encounters. What every such
planner takes for granted as input is a description of the surface to inspect together
with its orientation, and supplying that description is exactly the gap our pipeline
closes. Rather than relying on a hand-specified target, it provides oriented polygons that
carry their own surface normals, so that the inspection planner can be seeded directly
from the Explorer's semantic findings.

\section{Methodology}\label{sec:method}

The perception pipeline operates within a two-robot mission, summarized in
Fig.~\ref{fig:pipeline}. An \emph{Explorer} runs the STAGE exploration
planner~\cite{patel2024stage} and incrementally builds a globally consistent
point-cloud map of the mine using LiDAR-inertial odometry. Riding on top of this map,
the proposed pipeline performs onboard, open-set mineral perception. It segments
mineral deposits from the Explorer's RGB stream using a natural-language prompt,
projects the detections into the 3D map, clusters them into persistent regions of
interest, and for each region fits a plane and emits an oriented polygon. At the
end of the exploration mission the Explorer transmits two products: (i) the accumulated
point-cloud map (PCD), and (ii) the set of mineral polygons. A second
\emph{Inspector} robot, carrying an advanced sensing payload (e.g.\ a hyperspectral
camera), receives these products and uses the adaptive inspection
planner~\cite{viswanathan2025adaptive} to generate close-range viewpoints over each
polygon, oriented along its normal.


\subsection{Zero-shot semantic and panoptic segmentation}\label{sec:seg}
The semantic front-end is the \clipseg{} model,
which pairs a frozen CLIP backbone with a transformer decoder. 
image $I \in \mathbb{R}^{H\times W\times 3}$ and a set of $P$
natural-language prompts $\{t_1,\dots,t_P\}$ (e.g.\ \texttt{"white mineral"}), the model
produces, for each prompt $p$, a logit map that we pass through a sigmoid to obtain a
per-pixel confidence
\begin{equation}
  c_p(u,v) = \sigma\!\big(f_\theta(I, t_p)\big)(u,v) \in [0,1].
\end{equation}

A semantic label map is obtained by a thresholded argmax. We prepend a constant
``background'' channel equal to a confidence threshold $\tau$ to the $P$ confidence
maps and take the argmax over the resulting $P{+}1$ channels:
\begin{equation}
  L(u,v) = \arg\max_{p\in\{0,1,\dots,P\}}
           \big[\, \tau,\; c_1(u,v),\dots,c_P(u,v)\,\big],
  \label{eq:argmax}
\end{equation}
so that a pixel is assigned to mineral class $p$ only if $c_p(u,v) > \tau$ and $p$ is
the most confident class; otherwise $L(u,v)=0$ (background). This converts the soft
confidence maps into a hard, multi-class mask while suppressing low-confidence
responses. 

To separate distinct deposits that share the same mineral class, we extract a
panoptic (instance) labeling. For each class, external contours are computed on
its binary mask; contours whose area falls below a minimum $a_{\min}$ are discarded as
spurious, and each surviving contour is assigned a unique instance identifier, yielding
an instance map $S(u,v)$. The threshold $a_{\min}$ rejects small, flickering responses
that would otherwise generate unstable detections. 

\subsection{LiDAR projection and semantic point-cloud augmentation}\label{sec:proj}
The 2D labeling is grounded in 3D by projecting the time-synchronized LiDAR point
cloud onto the image. Image and cloud are paired by an approximate-time synchronizer.
A LiDAR point $\vct{X}=(x,y,z)$ in the camera frame is projected to a pixel
$(u,v)$ through the calibrated pinhole-plus-distortion model
\begin{equation}
  \tilde{\vct{p}} = K\,\big[\,R \mid \vct{t}\,\big]\,\tilde{\vct{X}},
  \qquad (u,v) = \pi\big(\tilde{\vct p}\big),
\end{equation}
where $K$ is the intrinsic matrix, $(R,\vct{t})$ the camera--LiDAR extrinsics, the lens
distortion coefficients are applied during projection, and $\pi(\cdot)$ denotes
perspective division, i.e.\ normalization by the homogeneous coordinate so that
$(u,v)=(\tilde p_1/\tilde p_3,\,\tilde p_2/\tilde p_3)$. Only points that fall in front of
the camera and within the image
bounds are retained. Each retained point inherits the attributes of the pixel it
projects to, namely the mineral confidence $c$, the semantic label $L$, the instance
identifier $S$, and the packed RGB color. The result is an augmented point
cloud whose per-point fields are
\begin{equation}
  \vct{q} = (x,\,y,\,z,\,c,\,L,\,S,\,\mathrm{rgb}).
\end{equation}

\subsection{Cross-view clustering and persistent regions of interest}\label{sec:cluster}
A global mapping node consumes the augmented clouds together with the robot's odometry
and condenses the stream of noisy, redundant, single-frame detections into a small,
stable set of mineral regions. This proceeds in three steps.

\emph{(i) Per-instance outlier rejection.} For each instance present in the incoming
cloud, the points are clustered with DBSCAN (radius $\varepsilon$,
minimum points $m$) and only the largest cluster is retained. This removes the thin
``spray'' of mislabeled points that arises when a segmentation mask straddles a depth
discontinuity and a few pixels project onto distant geometry.

\emph{(ii) Transformation to the world frame.} Using the synchronized odometry pose
$T_{wb}=(R_{wb},\vct{t}_{wb})$, the surviving points are transformed from the body
frame into the global map frame, $\vct{X}_w = R_{wb}\vct{X}_b + \vct{t}_{wb}$, so that
detections from different viewpoints become spatially comparable.

\emph{(iii) Cross-view association and fusion.} Each world-frame instance is wrapped in
a \emph{labeled-bbox} primitive that stores its constituent points and its mineral
class. A new primitive is merged into an existing global box of the same class
when their axis-aligned bounding boxes overlap, measured by the intersection-over-minimum
volume
\begin{equation}
  \mathrm{IoU}_{\cap\min}(A,B)=\frac{\mathrm{vol}(A\cap B)}{\min\!\big(\mathrm{vol}(A),\mathrm{vol}(B)\big)} > \rho,
\end{equation}
in which case the two point sets are concatenated; otherwise it is inserted as a new
bounding box. Because association keys on geometric overlap and semantic class rather than on
frame index, a deposit observed from many viewpoints accumulates into a single,
persistent bounding box that grows and stabilizes over time. Each persistent bounding box thus represents
one region of interest, and the set of bounding boxes is the clustered, deduplicated summary of
all mineralization seen during the mission. 

\subsection{Plane fitting and polygon generation}\label{sec:plane}
The final stage turns each persistent
region of interest into the oriented polygon that is handed to the Inspector.
Whereas the bounding box of Sec.~\ref{sec:cluster} captures where a deposit is,
inspection additionally requires how the surface is oriented and a bounded,
planar description of the area to be observed. We obtain both by fitting a plane to the mineral points and extracting the
minimum-area bounding rectangle of their projections onto that plane. 

\paragraph{Plane fitting} Let $\mathcal{P}=\{\vct{x}_i\}_{i=1}^{N}\subset\mathbb{R}^3$
be the mineral-labeled points accumulated inside a persistent bounding box. Mineralization on a
mine face is, to first order, a locally planar patch, so we estimate its dominant plane
\begin{equation}
  \pi:\;\; \vct{n}^{\!\top}\vct{x} + d = 0, \qquad \norm{\vct{n}}=1,
\end{equation}
with RANSAC. Repeatedly, three points are
sampled, the candidate plane is scored by the number of inliers within a distance
$\delta_\pi$, and the best hypothesis is refined by a least-squares fit to its inliers.
RANSAC is essential here because the accumulated cloud still contains off-surface
points from projection noise and from the rough mine wall behind the mineral vein. The
estimate yields the unit normal $\vct{n}$ and offset $d$; the normal is then oriented to
point towards observed free space (the side from which the Explorer viewed the face),
so that it defines a valid inspection viewing direction.

\paragraph{Projecting points and extracting the polygon}
The fitted plane is infinite, whereas the inspection target is finite and localized.
We bound it by projecting the mineral-labeled point cloud $\mathcal{P}$ directly onto
$\pi$ and fitting a minimum-area bounding rectangle to the projections.

Let $(\vct{e}_1,\vct{e}_2)$ be an orthonormal basis spanning $\pi$, obtained by
Gram--Schmidt from an arbitrary reference vector. Each mineral point is projected as
\begin{equation}
  \hat{\vct{x}}_i = \vct{x}_i - \bigl(\vct{n}^{\!\top}\vct{x}_i + d\bigr)\,\vct{n}
  \label{eq:proj}
\end{equation}
and embedded in the 2-D plane frame, centered at the projection centroid
$\bar{\vct{x}} = \tfrac{1}{N}\sum_i\hat{\vct{x}}_i$, as
\begin{equation}
  \vct{\xi}_i = \bigl(\vct{e}_1^{\!\top}(\hat{\vct{x}}_i-\bar{\vct{x}}),\;
                      \vct{e}_2^{\!\top}(\hat{\vct{x}}_i-\bar{\vct{x}})\bigr).
  \label{eq:embed}
\end{equation}
A minimum-area bounding rectangle is fitted to $\{\vct{\xi}_i\}$, yielding four
ordered corners $\{(c_{k,1},c_{k,2})\}_{k=1}^{4}$. Each corner is lifted back to
$\mathbb{R}^3$ as
\begin{equation}
  \vct{p}_k = \bar{\vct{x}} + c_{k,1}\,\vct{e}_1 + c_{k,2}\,\vct{e}_2,
  \quad k=1,\dots,4.
  \label{eq:lift}
\end{equation}
The ordered quadrilateral
\begin{equation}
  \Pi = \bigl(\vct{p}_1,\vct{p}_2,\vct{p}_3,\vct{p}_4\bigr)
  \label{eq:poly}
\end{equation}
together with $\vct{n}$ and $\bar{\vct{x}}$ is the mineral polygon.
Because the rectangle is fitted to the actual projections of the mineral points, it
tightly covers the full spatial support of the observed mineralization irrespective of
the plane's orientation relative to the coordinate axes. Large, elongated veins yield
large rectangles; compact pockets yield small ones, without any hand-tuned size
parameter.

\paragraph{From polygon to inspection pose} The polygon serves directly as input to the
Inspector's viewpoint planner~\cite{viswanathan2025adaptive}. A nominal observation pose
is obtained by standing off the centroid along the (outward) normal,
\begin{equation}
  \vct{v} = \bar{\vct p} + \eta\,\vct{n}, \qquad
  \text{heading}~=~-\vct{n},
\end{equation}
where $\eta$ is a payload-dependent stand-off distance; the planner then refines a full
set of coverage viewpoints over $\Pi$ subject to the robot's footprint and the tunnel
geometry. The complete message from Explorer to Inspector is the set of polygons
$\{(\Pi_j,\vct{n}_j,\bar{\vct p}_j)\}_j$ plus the shared PCD map, a description that is
orders of magnitude smaller than the raw semantic cloud yet sufficient to drive
close-range inspection.

\subsection{Graph-based organization of inspection targets}\label{sec:graph}
The polygons produced above are not transmitted as a flat list. Following the
semantically driven area-delimitation of~\cite{saucedo2026areas}, they are organized
into a three-level 3D scene graph. Each per-deposit polygon, together with its parent
bounding box, forms a leaf node; spatially coherent leaves are grouped into
\emph{mineral-area} nodes, so that each area collects several nearby deposits while every
deposit retains its own polygon; and all areas attach to a single root node representing
the mine. Two properties follow. First, ranking the areas by the density of deposits they
contain yields a priority ordering, so that the Inspector is dispatched to the most
densely mineralize areas first. Second, the same partitioning decomposes the target set
into spatially coherent subsets that can be assigned to separate Inspector robots for
parallel coverage. The experiments below use a single Inspector, but this organization
makes the hand-over directly extensible to a fleet.

\section{Experimental Validation}\label{sec:experiments}

The pipeline was validated in two qualitatively different subterranean environments and
on two mineralization types. 
The first is the Subterranean (SubT) test facility at
Lule{\aa} University of Technology, where the target was the irregular iron-vein
formations embedded in the tunnel walls. The second is an active, operator-owned
magnesite mine (Grecian Magnesite, Koutizi), where the target was the magnesite
mineralization distributed along the drift. In both cases a ground robot equipped with
a calibrated RGB camera and a 3D LiDAR ran the perception stack onboard, with depth
obtained by projecting the LiDAR cloud into the image and odometry provided by
LiDAR-inertial estimation. The only deployment-specific change between sites was the
natural-language prompt; no weights were fine-tuned and no mine-specific data were
collected. Unless noted otherwise, the segmentation confidence threshold, minimum
contour area, clustering radius, and association overlap were held fixed across sites.

\begin{figure}[!h]
  \centering
  \includegraphics[width=\linewidth]{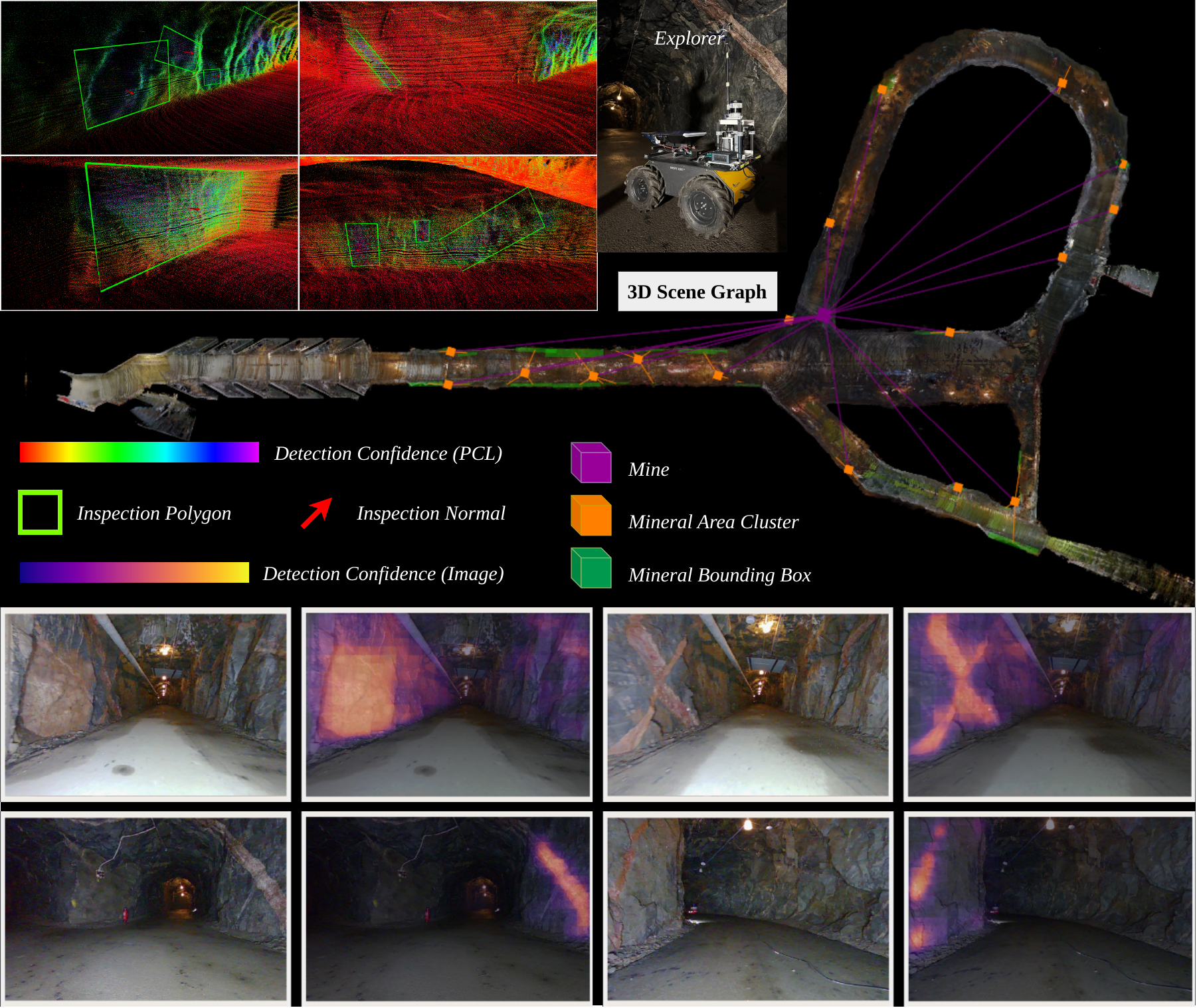}
  \caption{\textbf{Iron-vein results (LTU SubT facility).} \emph{Right/center}: the
  reconstructed mine point cloud overlaid with the onboard-generated 3D scene graph, in
  which a root \emph{Mine} node (magenta) connects to the \emph{mineral area clusters}
  (orange), each grouping the per-deposit \emph{mineral bounding boxes} (green). \emph{Top-left insets}: the oriented
  inspection polygons (green) and their surface normals (red), fitted to representative
  deposits, 
  the cloud
  is colored by per-point detection confidence. \emph{Bottom}: eight representative onboard frames with the per-image
  \clipseg{} detection-confidence heat-maps from which the mineral masks are extracted.
  }
  \vspace{3mm}
  \label{fig:iron}
\end{figure}
\begin{figure}[!h]
  \centering
  \includegraphics[width=\linewidth]{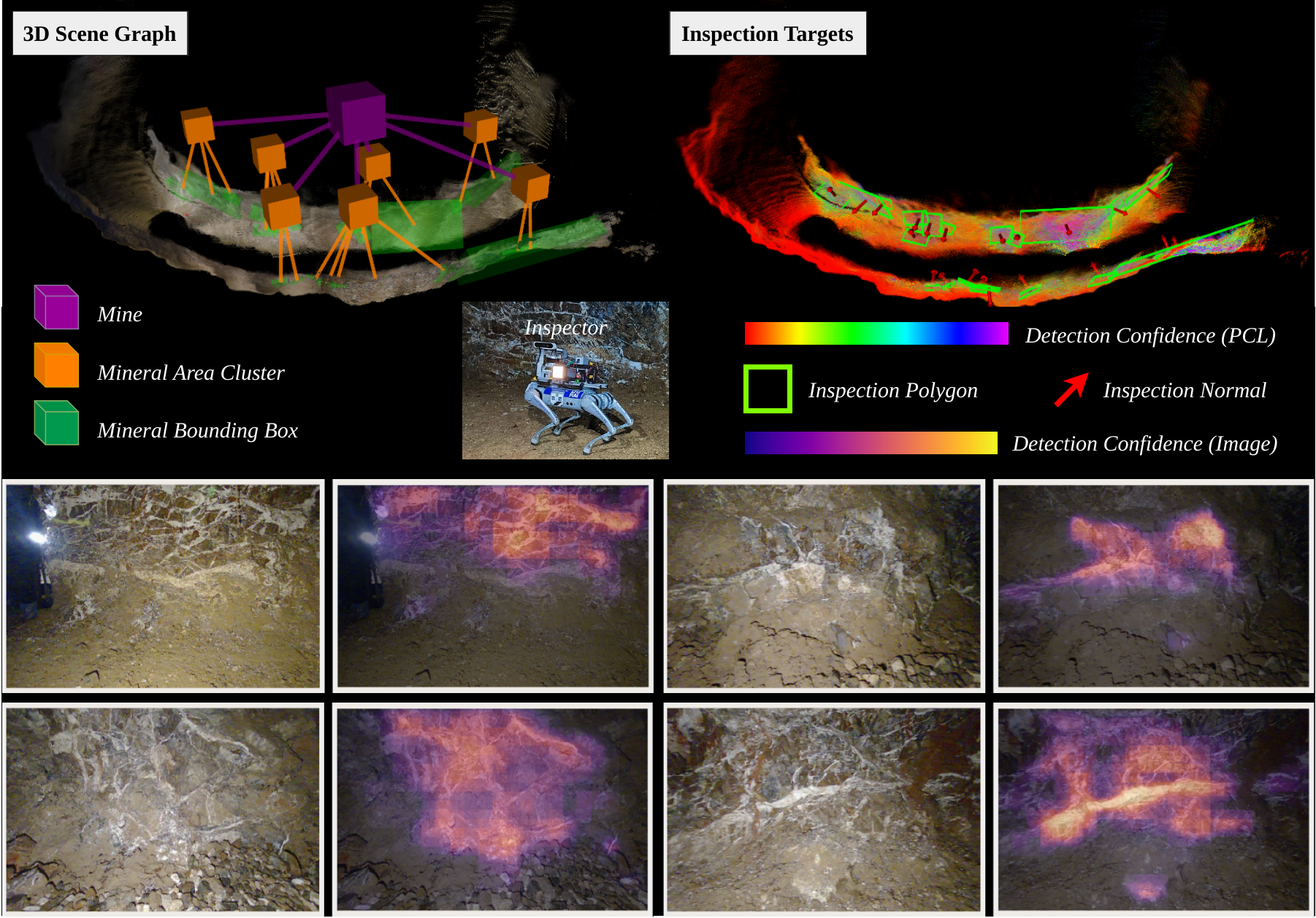}
  \caption{\textbf{Magnesite results (Grecian Magnesite, Koutizi).} 
  \emph{Top-left}: the 3D scene graph linking the
  root \emph{Mine} node (magenta) to its \emph{mineral area clusters} (orange) and
  \emph{mineral bounding boxes} (green). \emph{Top-right}: the resulting
  \emph{inspection targets}, i.e.\ the oriented polygons (green) and surface normals (red)
  over the confidence-colored point cloud. 
  \emph{Bottom}: per-image \clipseg{} detection-confidence
  heat-maps. 
  }
  \label{fig:magnesite}
\end{figure}
\begin{figure}[htbp]
  \centering
  \begin{subfigure}{0.49\linewidth}
    \includegraphics[width=\linewidth]{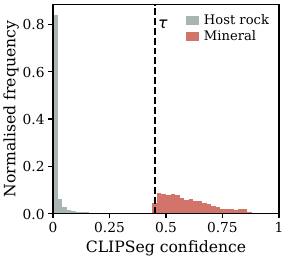}
    \caption{Confidence Spread}\label{fig:quant_conf}
  \end{subfigure}\hfill
  \begin{subfigure}{0.49\linewidth}
    \includegraphics[width=\linewidth]{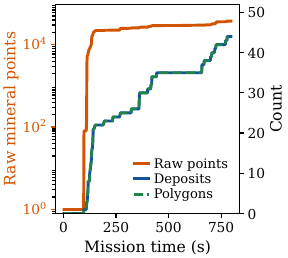}
    \caption{Map condensation}\label{fig:quant_cond}
  \end{subfigure}
  \vspace{6mm}

  \begin{subfigure}{0.49\linewidth}
    \includegraphics[width=\linewidth]{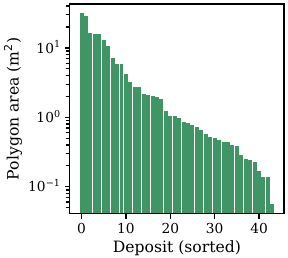}
    \caption{Target extents}\label{fig:quant_area}
  \end{subfigure}\hfill
  \begin{subfigure}{0.49\linewidth}
    \includegraphics[width=\linewidth]{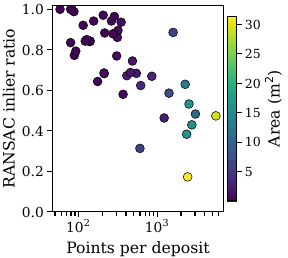}
    \caption{Plane-fit quality}\label{fig:quant_qual}
  \end{subfigure}
  \vspace{5mm}
  \caption{\textbf{Quantitative analysis.}
  \textbf{(a)}~Per-point \clipseg{} detection confidence cleanly separates the detected
  mineral ($\bar c\!\approx\!0.60$) from the host rock ($\bar c\!\approx\!0.02$), with the
  segmentation threshold $\tau$ (dashed) lying in the valley between the two modes.
  \textbf{(b)}~Over the run the ${\sim}38$k raw mineral points (orange) are condensed by
  the cross-view clustering of Sec.~\ref{sec:cluster} into a few dozen persistent deposits
  and inspection polygons (blue/green), a handover roughly two orders of magnitude smaller
  than the raw semantic cloud.
  \textbf{(c)}~Distribution of inspection-polygon areas across deposits, spanning compact
  pockets to larger vein faces.
  \textbf{(d)}~RANSAC plane-fit inlier ratio versus deposit size, color corresponds to polygon area,
  compact deposits are well approximated by a single plane, while the largest deposits are
  less planar and are better described by multiple polygons.}
  \label{fig:quant}
\end{figure}

Figures~\ref{fig:iron} and~\ref{fig:magnesite} summarize the end-to-end pipeline output at
the two sites, while Fig.~\ref{fig:quant} presents quantitative metrics for the SubT iron-vein run. The per-image detection-confidence maps (bottom rows) show that \clipseg{} consistently
highlights the mineralize regions and suppresses the surrounding host rock, but the two
sites stress the front-end in different ways. At the SubT facility the iron veins appear
as thin, high-contrast lineations crossing otherwise uniform tunnel walls, and the
confidence maps resolve them as sharp, well-localized ridges that follow the vein
geometry. In the magnesite mine the mineralization is broader and more diffuse,
distributed as irregular bright patches over visually rougher rock; the same model still
concentrates its response on the magnesite and rejects the background, although the
resulting confidence fields are correspondingly less crisp. Both behaviors persist
despite irregular surface texture, strong specular reflections, cast shadows from the
robot's own lighting, and airborne dust. Crucially, the entire transfer between the two
mineral types is achieved by changing only the text prompt; no weights are retrained and
no site-specific data are collected, which is the central practical advantage of the
zero-shot formulation in a setting where labeled data are unavailable.

Projecting these masks onto the LiDAR and accumulating them across the trajectory yields a
globally consistent, semantically annotated point cloud, color-coded by detection
confidence over the reconstructed tunnel geometry. 
The
per-instance outlier rejection and cross-view association of Sec.~\ref{sec:cluster} reduce
the thousands of noisy per-frame mineral points into a handful of persistent bounding
boxes, each corresponding to one physical deposit. These boxes are then grouped into
mineral-area clusters and attached to a single root node, forming the three-level 3D scene
graph rendered in the upper panels. The hierarchy is informative in itself: areas that
collect many nearby deposits stand out as the densest mineralization and become the first
inspection priorities, while isolated detections are demoted automatically. The same
structure is recovered at both sites despite their very different drift geometries, which
indicates that the condensation and grouping do not depend on a particular tunnel layout.

For every persistent region, RANSAC plane fitting and minimum-area rectangle extraction
produce an oriented inspection polygon and its surface normal, shown as the green polygons
and red arrows in the inspection-target views. The polygons track the spatial support of
each deposit tightly, from large elongated iron veins down to compact magnesite pockets,
and the normals consistently point into the free space from which the deposit was
observed, so each target encodes both \emph{where} to look and \emph{from which
direction}. The complete handover is just this set of polygons together with the shared
PCD map, a description orders of magnitude smaller than the raw semantic cloud and
therefore transmissible over a degraded underground link. 

\section{Conclusions}\label{sec:conclusion}
We presented an onboard, open-set perception pipeline that serves as the interface
between two heterogeneous robots in an autonomous underground mining mission. The
pipeline detects mineral deposits from natural-language prompts using zero-shot
vision-language segmentation, grounds them in 3D through LiDAR projection, condenses
redundant multi-view evidence into persistent regions of interest by clustering and
cross-view association, and converts each region into an oriented polygon
by fitting a plane to its mineral points, projecting those points onto the fitted plane,
and extracting their minimum-area bounding rectangle. The resulting polygons, organized as a 3D scene graph of inspection
targets that supports density-based prioritization,
together with the shared map, form a compact and actionable description that an Explorer
can transmit to an Inspector to seed close-range viewpoint planning. Field validation in a subterranean test facility and in
an active magnesite mine, on two mineralization types, demonstrated robust, training-free
operation under realistic underground conditions. 

\section*{Acknowledgment}

During the preparation of this work, the author(s) used the Claude model developed by Anthropic, in order to correct grammar and refine the wording of the manuscript.


\bibliographystyle{ieeetr}
\bibliography{references}

\end{document}